\documentclass{article}
\usepackage{arxiv}
\usepackage{array}
\usepackage[utf8]{inputenc} 
\usepackage[T1]{fontenc}    
\usepackage{hyperref}       
\usepackage{url}            
\usepackage{booktabs}       
\usepackage{amsmath}
\usepackage{amssymb}
\usepackage{amsfonts}       
\usepackage{nicefrac}       
\usepackage{microtype}      
\usepackage{caption}
\usepackage{subcaption}
\usepackage{algorithm, algcompatible, algpseudocode, float}
\usepackage{setspace}
\usepackage{graphicx}
\usepackage{ctable}
\usepackage{bbm}
\usepackage{dsfont}
\usepackage{appendix}
\DeclareMathOperator*{\argmax}{argmax}

\title{Knowledge Distillation for BERT Unsupervised Domain Adaptation}

\author{Minho Ryu\thanks{This work was done while the author was a graduate student in the Department of Industrial Engineering, Hanyang University.}\\
Vision AI Labs \\
SK Telecom \\
{\tt ryumin93@sktbrain.com}\\[4pt]
\And
Kichun Lee\\
Department of Industrial Engineering \\ 
Hanyang University \\
{\tt skylee@hanyang.ac.kr}\\[4pt]
}

\begin{document}

\maketitle
\begin{abstract}
A pre-trained language model, BERT, has brought significant performance improvements across a range of natural language processing tasks. Since the model is trained on a large corpus of diverse topics, it shows robust performance for domain shift problems in which data distributions at training (source data) and testing (target data) differ while sharing similarities. Despite its great improvements compared to previous models, it still suffers from performance degradation due to domain shifts. To mitigate such problems, we propose a simple but effective unsupervised domain adaptation method, \emph{adversarial adaptation with distillation} (AAD), which combines the adversarial discriminative domain adaptation (ADDA) framework with knowledge distillation. We evaluate our approach in the task of cross-domain sentiment classification on 30 domain pairs, advancing the state-of-the-art performance for unsupervised domain adaptation in text sentiment classification.
\end{abstract}

\section{Introduction}
The cost of creating labeled data for a new machine learning task is often a major obstacle to the application of machine learning algorithms. In particular, the obstacle is more restrictive for deep learning architectures that require huge datasets to learn a good representation.
Even if enough data are available for a particular problem, performance may degrade due to distribution changes in training, testing, and actual service.

Domain adaptation is a way for machine learning models trained on \textit{source} domain data to maintain good performance on \textit{target} domain data.
In domain adaptation methods, semi-supervised methods require a small amount of labeling in a target domain while unsupervised methods do not.
Although semi-supervised methods may provide better performance, unsupervised domain adaptation methods are more often noticeable and attractive because of the high cost of data annotation depending on the new domain.

With the development of deep neural networks, unsupervised domain adaptation methods have focused on learning to map source and target data into a common feature space. This is usually accomplished by optimizing the representation to minimize some measure of domain shifts such as maximum mean discrepancy~\citep{Tzeng2014DDC} or correlation distances~\citep{Sun2016CORAL, Sun2016DeepCORAL}. Particularly, adversarial domain adaptation methods have become more popular in recent years, seeking to minimize domain discrepancy distance through an adversarial objective~\citep{Ganin2016DANN, Tzeng2017AdversarialDD}.

However, these methods appear to be unfavorable when applied to large-scale and pre-trained language models such as BERT~\citep{Devlin2018BERT}.  Pre-trained language models~\citep{Peters2018DeepCWR, Radford2018ImprovingLU, Devlin2018BERT, Yang2019XLNet} have brought tremendous performance improvements in numerous natural language processing (NLP) tasks. With respect to domain shift issues, showing robust performance and outperforming existing models without domain adaptation, they still suffer from performance degradation due to domain shifts.

In this paper, we propose a novel adversarial domain adaptation method for pre-trained language models, called \emph{adversarial adaptation with distillation} (AAD). This work is done on top of the framework, called adversarial discriminative domain adaptation (ADDA), proposed by Tzeng \textit{et al}.~\citep{Tzeng2017AdversarialDD}. We observe that a catastrophic forgetting~\citep{kirkpatrick2016overcoming} occurs when the ADDA framework is applied to the BERT model as opposed to when applied to deep convolutional neural networks.
In ADDA, the fine-tuned source model is used as an initialization to prevent the target model from learning degenerate solutions because the target model is trained without label information. Unfortunately, this method alone does not prevent a catastrophic forgetting in BERT, resulting in random classification performance. To overcome this problem, we adopt the knowledge distillation method~\citep{Hinton2015KD}, which is mainly used to improve the performance of a smaller model by transferring knowledge from a large model. We found that this method can serve as a regularization to maintain the information learned by the source data while enabling the resulting model to be domain adaptive and to avoid overfitting.

\begin{figure*}[t]
     \centering
     \includegraphics[width=\linewidth]{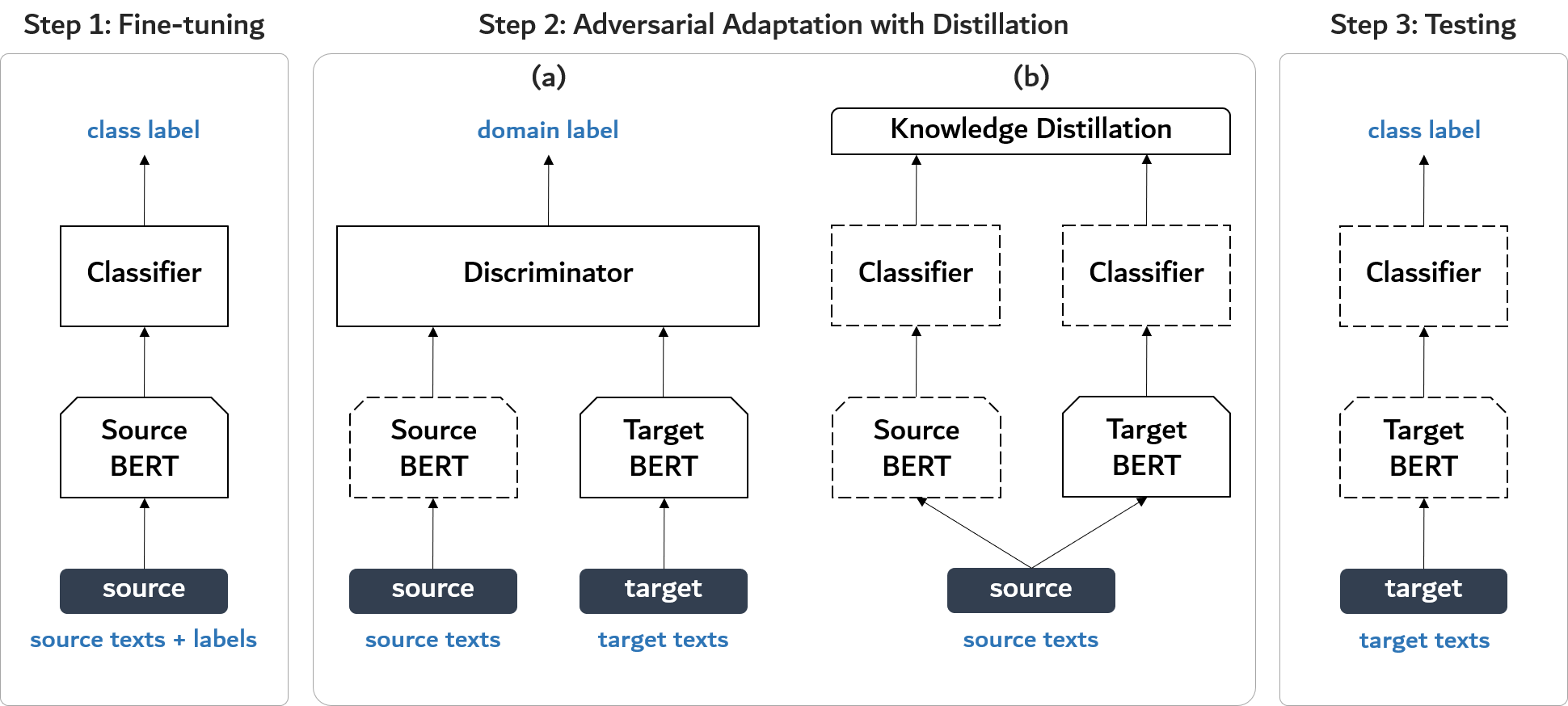}
     \caption{An overview of Adversarial Adaptation with Distillation. In Step 1, we first fine-tune a source BERT and a classifier on source labeled data. Next, in Step 2, after initializing the target BERT with weights of the fine-tuned source BERT, we perform adversarial learning and knowledge distillation simultaneously on the target BERT. Dashed line indicates fixed model parameters. Finally, in Step 3, the knowledge-distilled target BERT and the classifier predict class labels}
    \label{fig:fig1}
\end{figure*}

\section{Related Work}

\subsection{Unsupervised Domain Adaptation}

Recently a large number of unsupervised domain adaptation methods have been studied. We present details of the studies that are most relevant to our paper. Recent studies have focused on transferring deep neural network representations learned from labeled source data to unlabeled target data. 

Deep Domain Confusion (DDC)~\citep{Tzeng2014DDC} introduces an adaptation layer to minimize Maximum Mean Discrepancy (MMD) in addition to classification loss on source data while the 
Deep Adaptation Network (DAN)~\citep{Long2015DAN} applies multiple kernels to multiple layers. The deep Correlation Alignment (deep
CORAL)~\citep{Sun2016DeepCORAL} minimizes the difference in second-order statistics between the source and target representations. 

More recently, adversarial methods to minimize domain shifts have received much attention. The Domain Adversarial Neural Network (DANN)~\citep{Ganin2016DANN} introduces a domain binary classification with a gradient reversal layer to train in the presence of domain confusion. Other studies have explored generative methods using Generative Adversarial Networks (GANs)~\citep{Goodfellow2014GAN}. 
A coupled generative adversarial network (CoGAN)~\citep{Liu2016CoGAN} learns a joint distribution from the source and the target data with weight sharing constraints. 
Cycle-Consistent Adversarial Domain Adaptation (CyCADA)~\citep{Hoffman2018CyCADA} uses cycle and semantic consistency for multi-level adaptation.

ADDA~\citep{Tzeng2017AdversarialDD} was proposed as an adversarial framework that includes discriminative modeling, untied weight sharing, and a GAN-based loss. The source encoder is first trained with labeled source data and the weights are copied to the target encoder. Then, the target encoder and discriminator are alternately optimized in a two-player game like the original GAN setting. The discriminator learns to distinguish the target representations from the source representations while the encoder learns to trick the discriminator. Chadha \textit{et al}.~\citep{Chadha2018ImprovingAD} improved the ADDA framework by modifying the discriminator to jointly predict the source labels and distinguish inputs from the target domain as semi-supervised GANs~\citep{Kumar2017SSGAN}. Our study is similar to the work by Chadha \textit{et al}. (2018) in that it also uses source information in the adversarial adaptation step. However, the difference is that the use of knowledge distillation, rather than the direct use of the source label, is the means to employ the source information in the network.

Besides, several unsupervised domain adaptation methods designed for NLP have also been proposed. Structural Corresponding Learning (SCL)~\citep{Blitzer2006SCL} identifies correspondences among features from different domains by modeling their correlations with \emph{pivot} features. Neural SCL~\citep{Ziser2017NeuralSCL} incorporates ideas of SCL and autoencoder neural networks. The Pivot Based Language Model (PBLM)~\citep{Ziser2018PBLM} also combines the pivot-based idea of SCL with neural network based language modeling.

\subsection{Knowledge Distillation} 
Knowledge Distillation~\citep{Hinton2015KD} (KD) is originally a model compression technique that aims to train a compact model (student) so that the knowledge of a well-trained larger model (teacher) is transferred to the student model. KD can be formulated by minimizing the following objective function

\begin{align}
    \mathcal{L}_{KD} = t^2\times \sum_{k}{-\textrm{softmax}(\boldsymbol{z}_k^T/t)
    \times \log(\textrm{softmax}(\boldsymbol{z}_k^S/t))}
    \label{eq1}
\end{align}

where $\boldsymbol{z}^S$ and $\boldsymbol{z}^T$ are the logits predicted by the student and the teacher, respectively, and temperature value $t$ controls the degree of knowledge transfer. Equation~\ref{eq1} can be derived from the Kullback-Leibler (KL) divergence\footnote[1]{Given probability distributions $p$ and $q$, KL divergence of $q$ from $p$ is defined to be $KL(p \parallel q)=-\sum_j{p_j\log(q_j/p_j)}$} of the predicted distribution by the teacher from the predicted distribution by the student since the teacher model is fixed during training.

In supervised learning, the standard training objective is to minimize the cross-entropy between the distribution of the model's predicted probability and that of one-hot-encoded labels' true probability. However, this objective is prone to result in overfitting with repeated training epochs.
Since a larger value for $t$ produces a softer probability distribution, knowledge distillation can mitigate this problem when incorporated with domain adaptation methods.

\subsection{Bidirectional Encoder Representations from Transformers}
BERT is a self-supervised approach for pre-training a deep transformer encoder~\citep{Vaswani2017Transformer}. The BERT model is trained on a large corpus using masked language modeling and next sentence prediction.
It has shown strong performance gains in many NLP tasks, and several variants have been proposed such as spanBERT~\citep{Joshi2019SpanBERT}, distilBERT~\citep{sanh2019distilbert}, and RoBERTa~\citep{Liu2019Roberta}. In this experiments, we use BERT, distilBERT and RoBERTa to evaluate our approach.

\section{Adversarial Adaptation with Distillation}
In this section, we introduce our unsupervised domain adaptation method, \emph{adversarial adaptation with distillation}, which combines the ADDA framework~\citep{Tzeng2017AdversarialDD} with knowledge distillation. We illustrate the proposed method in Figure \ref{fig:fig1}.

Let us assume labeled source texts are given, $\mathbf{X}_S=\{(\boldsymbol{x}_s^i)\}_{i=0}^{N_s}$, $\mathbf{y}_S=\{(y_s^i)\}_{i=0}^{N_s}$ with $(\boldsymbol{x}_s, y_s) \sim (\mathbb{X}_S, \mathbb{Y}_S)$, and unlabeled target texts are also given, $\mathbf{X}_T=\{(\boldsymbol{x}_t^i)\}_{i=0}^{N_t}$ with $\boldsymbol{x}_t \sim \mathbb{X}_T$. 
We also assume that the target data share the identical label space as the source data. The source encoder is represented by a function $E_s(\boldsymbol{x})$ where $\boldsymbol{x}$ is the input to the network, and likewise, $E_t(\boldsymbol{x})$ represents the target encoder. 
In addition, let us represent $C$ as a classifier function that maps the source encoder output to class probabilities and $D$ as a discriminator function that maps the encoder output (of either source or target) to domain probabilities. 
In unsupervised domain adaptation, the goal is to have better performance on target data by learning to minimize the distance between the representation of source data and that of target data without access to the target labels. 
Our proposed method consists of the following three steps: training the source encoder and the classifier on the source data, adapting the target encoder to align its representation with the source representation through both adversarial training and distillation, and finally inferring on the target data with the adapted target encoder and the trained classifier. 

\subsection{Step 1: Fine-tune the source encoder and the classifier}
\label{section3.2:step1}
With access to the labeled source data, we first fine-tune the source encoder $E_s$ and the classifier $C$ on $\mathbf{X}_S$ and $\mathbf{y}_S$ using standard cross-entropy loss:

\begin{align}
    \min_{E_s, C}\mathcal{L}_{S}(\mathbf{X}_S, \mathbf{y}_S)=\mathbb{E}_{(\boldsymbol{x}_s, y_s) \sim (\mathbb{X}_S, \mathbb{Y}_S)}-\sum_{k=1}^K \mathds{1}_{[k=y_s]} \log{C(E_s(\boldsymbol{x}_s))}
    \label{eq2}
\end{align}
where $K$ is the number of classes.
Then, after initializing the target-encoder parameters with the fine-tuned source-encoder parameters, we freeze the source-encoder parameters and the classifier.

\subsection{Step 2: Adapt the target encoder via adversarial adaptation with distillation}
\label{section3.2:step2}
In this step, we train the target encoder and the discriminator alternately in the original GAN setting as in the ADDA framework. This can be formulated by the following unconstrained optimization as in Step 2-(a) of Figure~\ref{fig:fig1}:

\begin{align}
\begin{split}
    &\min_{D}~\mathcal{L}_{dis}(\mathbf{X}_S, \mathbf{X}_T)=\mathbb{E}_{\boldsymbol{x}_s \sim \mathbb{X}_S}-\log{D(E_s(\boldsymbol{x}_s))}+\mathbb{E}_{\boldsymbol{x}_t \sim \mathbb{X}_T}-\log{(1 - D(E_t(\boldsymbol{x}_t)))}, \\
    & \min_{E_t}~\mathcal{L}_{gen}(\mathbf{X}_T)=\mathbb{E}_{\boldsymbol{x}_t \sim \mathbb{X}_T}-\log{D(E_t(\boldsymbol{x}_t))}.
    \label{eq3}
\end{split}
\end{align}

Since it has the untied weights from the source encoder, the target encoder is allowed to have more flexibility to learn specific domain features. However, the formulation easily leads to catastrophic forgetting, resulting in random classification performance due to the inaccessibility to class labels and the dissimilarity to the original task. 

In order to enhance the stability of the adversarial training, one can think of using source labels directly as a supervised learning approach. However, this can cause the model to overfit the source domain data while possibly preventing a mode collapse in the adversarial adaptation. Knowledge distillation~\citep{Hinton2015KD}, on the other hand, can provide the model with both flexibility for adversarial adaptation and the ability to retain class information with a large temperature value $t$. 
Therefore, we introduce knowledge distillation loss as in Step 2-(b) of Figure~\ref{fig:fig1}:

\begin{align}
    \mathcal{L}_{KD}(\mathbf{X}_S)= t^2 \times \mathbb{E}_{\boldsymbol{x}_s \sim \mathbb{X}_S}\sum_{k=1}^K {-\textrm{softmax}(\boldsymbol{z}_k^S/t)\times\log(\textrm{softmax}(\boldsymbol{z}_k^T/t))}
\label{eq4}
\end{align}
$\text{where}~\boldsymbol{z}^S = C(E_s(\boldsymbol{x}_s)),~\boldsymbol{z}^T = C(E_t(\boldsymbol{x}_s))$. Thus, the final objective function for training target encoder becomes:

\begin{align}
    \min_{E_t}~\mathcal{L}_{T}(\mathbf{X}_S,  \mathbf{X}_T)=\mathcal{L}_{gen}(\mathbf{X}_T)+\mathcal{L}_{KD}(\mathbf{X}_S).
    \label{eq5}
\end{align}
Finally, the second objective function in equation~(\ref{eq3}) is replaced with equation~(\ref{eq5}). Then, the discriminator and the target encoder are trained by alternately minimizing objective functions.

\subsection{Step 3: Test the target encoder on the target data}
We can now test the target encoder on the target data. As illustrated in Step 3 of Figure \ref{fig:fig1}, we use the fine-tuned classifier for inference, obtaining the prediction as follows:

\begin{equation}
    \hat{y}_t = \argmax~C(E_t(\boldsymbol{x}_t)).
\end{equation}

\begin{table*}[t!]
\centering
\resizebox{10cm}{!}{
\begin{tabular}{c || lllll}
\specialrule{.1em}{.05em}{.05em}
Source $\rightarrow$ Target & Baseline & DDC & DANN & deep CORAL & AAD~(Ours) \\
\hline
B $\rightarrow$ D & \textbf{86.6} & 84.7\small{$\pm$1.9} & 85.5\small{$\pm$0.8} & 85.0\small{$\pm$2.3} & 86.5\small{$\pm$0.2} \\
B $\rightarrow$ E & 85.7 & 84.9\small{$\pm$2.5} & 84.3\small{$\pm$1.6} & 86.6\small{$\pm$0.7} & \textbf{86.7}$^*$\small{$\pm$0.2} \\
B $\rightarrow$ K & \textbf{88.4} & 87.0\small{$\pm$0.7} & 86.3\small{$\pm$2.0} & 87.6\small{$\pm$0.6} & 88.3\small{$\pm$0.3} \\
B $\rightarrow$ A & 84.8 & 82.8\small{$\pm$2.6} & 83.6\small{$\pm$1.2} & 84.3\small{$\pm$1.3} & \textbf{85.9}$^*$\small{$\pm$0.3} \\
B $\rightarrow$ I & \textbf{83.4} & 80.8\small{$\pm$3.5} & 82.3\small{$\pm$0.9} & 81.9\small{$\pm$0.9} & 82.4\small{$\pm$0.5} \\
D $\rightarrow$ B & 85.0 & 84.8\small{$\pm$1.0} & 83.7\small{$\pm$3.1} & 84.8\small{$\pm$1.6} & \textbf{87.0}$^*$\small{$\pm$0.2} \\
D $\rightarrow$ E & 83.8 & 83.2\small{$\pm$1.6} & 81.5\small{$\pm$1.9} & 84.5\small{$\pm$1.1} & \textbf{85.6}$^*$\small{$\pm$0.3} \\
D $\rightarrow$ K & 85.3 & 85.6\small{$\pm$0.6} & 85.7\small{$\pm$0.6} & 86.4$^*$\small{$\pm$0.7} & \textbf{86.7}$^*$\small{$\pm$0.5} \\
D $\rightarrow$ A & 81.0 & 81.9\small{$\pm$1.6} & 82.0\small{$\pm$1.4} & 83.2$^*$\small{$\pm$1.7} & \textbf{84.2}$^*$\small{$\pm$0.2} \\
D $\rightarrow$ I & 82.3 & 82.8\small{$\pm$1.6} & 81.6\small{$\pm$2.4} & 82.5\small{$\pm$1.7} & \textbf{82.9}$^*$\small{$\pm$0.1} \\
E $\rightarrow$ B & 85.0 & 84.0\small{$\pm$0.8} & 82.9\small{$\pm$1.0} & 84.4\small{$\pm$0.5} & \textbf{85.1}\small{$\pm$0.3} \\
E $\rightarrow$ D & 84.4 & 83.7\small{$\pm$0.8} & 81.7\small{$\pm$2.5} & 83.5\small{$\pm$0.7} & \textbf{84.6}\small{$\pm$0.3} \\
E $\rightarrow$ K & 90.6 & 90.1\small{$\pm$0.7} & 88.9\small{$\pm$0.6} & 88.9\small{$\pm$1.0} & \textbf{90.9}$^*$\small{$\pm$0.2} \\
E $\rightarrow$ A & 84.3 & 85.9$^*$\small{$\pm$1.1} & 85.0\small{$\pm$1.2} & 85.9$^*$\small{$\pm$1.0} & \textbf{86.4}$^*$\small{$\pm$0.3} \\
E $\rightarrow$ I & 79.1 & 80.0\small{$\pm$0.9} & 79.4\small{$\pm$1.2} & 80.2\small{$\pm$0.9} & \textbf{81.2}$^*$\small{$\pm$0.5} \\
K $\rightarrow$ B & \textbf{84.9} & 80.5\small{$\pm$4.4} & 82.6\small{$\pm$0.9} & 83.0\small{$\pm$1.7} & 84.4\small{$\pm$1.8} \\
K $\rightarrow$ D & 83.1 & 81.7\small{$\pm$1.5} & 82.5\small{$\pm$1.2} & 82.5\small{$\pm$2.0} & \textbf{83.7}$^*$\small{$\pm$0.3} \\
K $\rightarrow$ E & \textbf{88.1} & 86.8\small{$\pm$0.9} & 86.8\small{$\pm$1.3} & 87.7\small{$\pm$0.3} & \textbf{88.1}\small{$\pm$0.7} \\
K $\rightarrow$ A & 80.4 & 81.0\small{$\pm$2.4} & 83.1$^*$\small{$\pm$1.8} & 82.8\small{$\pm$2.0} & \textbf{85.9}$^*$\small{$\pm$0.5} \\
K $\rightarrow$ I & 80.2 & 78.3\small{$\pm$1.6} & 77.7\small{$\pm$1.4} & 79.4\small{$\pm$1.7} & \textbf{80.6}\small{$\pm$1.0} \\
A $\rightarrow$ B & 77.2 & 78.3\small{$\pm$1.7} & 77.1\small{$\pm$3.8} & 79.5\small{$\pm$2.8} & \textbf{80.9}$^*$\small{$\pm$0.7} \\
A $\rightarrow$ D & 77.7 & 77.8\small{$\pm$1.2} & 77.9\small{$\pm$1.3} & 78.7\small{$\pm$2.6} & \textbf{78.9}\small{$\pm$1.1} \\
A $\rightarrow$ E & 84.3 & 84.3\small{$\pm$1.0} & 83.9\small{$\pm$1.4} & 83.5\small{$\pm$1.8} & \textbf{85.5}$^*$\small{$\pm$0.4} \\
A $\rightarrow$ K & 85.0 & 84.6\small{$\pm$1.2} & 82.5\small{$\pm$2.1} & 85.2\small{$\pm$1.6} & \textbf{87.5}$^*$\small{$\pm$0.4} \\
A $\rightarrow$ I & 71.2 & 73.8\small{$\pm$2.9} & 75.2$^*$\small{$\pm$1.9} & \textbf{76.5}$^*$\small{$\pm$3.2} & 75.3$^*$\small{$\pm$1.6} \\
I $\rightarrow$ B & 84.5 & 83.0\small{$\pm$2.3} & 82.4\small{$\pm$2.4} & 84.8\small{$\pm$1.4} & \textbf{86.6}$^*$\small{$\pm$0.2} \\
I $\rightarrow$ D & 84.8 & 84.5\small{$\pm$1.3} & 84.0\small{$\pm$2.5} & 84.7\small{$\pm$1.8} & \textbf{85.9}$^*$\small{$\pm$0.2} \\
I $\rightarrow$ E & 82.0 & 83.3\small{$\pm$1.5} & 83.9$^*$\small{$\pm$0.8} & 84.9$^*$\small{$\pm$0.7} & \textbf{86.3}$^*$\small{$\pm$0.3} \\
I $\rightarrow$ K & 85.2 & 84.6\small{$\pm$0.7} & 84.6\small{$\pm$1.1} & 86.2\small{$\pm$1.2} & \textbf{87.4}$^*$\small{$\pm$0.1} \\
I $\rightarrow$ A & 82.0 & 83.4$^*$\small{$\pm$1.0} & 82.1\small{$\pm$1.7} & 83.8\small{$\pm$1.5} & \textbf{84.9}$^*$\small{$\pm$0.4} \\
\hline
Average & 83.3 & 82.9~(2$^\dagger$) & 82.7~(3$^\dagger$) & 83.8~(5$^\dagger$) & \textbf{84.9}~(21$^\dagger$) \\
\specialrule{.1em}{.05em}{.05em}
\end{tabular}
}
\caption{Sentiment classification accuracy with $\text{BERT}_\text{BASE}$ and other tested models including the proposed model ADD, on 30 cross-domain sentiment classification tasks is shown.  The asterisk, $^*$, denotes a value greater than the baseline with a significance level of 0.05, and $^\dagger$ represents the number of values significantly greater than the baseline.}
\label{tab:BERT_results}
\end{table*}

\begin{table*}[t!]
\centering
\resizebox{14cm}{!}{
\begin{tabular}{c || llllllll}
\specialrule{.1em}{.05em}{.05em}
Source $\rightarrow$ Target & Baseline & Supervised & $t=1$ & $t=2$ & $t=5$ & $t=10$ & $t=20$ & $t=50$ \\
\hline
B $\rightarrow$ D & \textbf{86.6} & 84.3\small{$\pm$2.6} & 86.2\small{$\pm$0.7} & 86.0\small{$\pm$1.2} & 86.1\small{$\pm$1.2} & 85.9\small{$\pm$1.1} & 86.5\small{$\pm$0.2} & 86.0\small{$\pm$1.1} \\
B $\rightarrow$ E & 85.7 & 85.1\small{$\pm$1.2} & 85.3\small{$\pm$0.7} & 85.9\small{$\pm$0.4} & 86.7$^*$\small{$\pm$0.5} & 86.3\small{$\pm$0.4} & \textbf{86.7}$^*$\small{$\pm$0.2} & 86.4\small{$\pm$0.5} \\
B $\rightarrow$ K & \textbf{88.4} & 88.0\small{$\pm$0.5} & 87.9\small{$\pm$0.6} & 88.1\small{$\pm$0.6} & 88.1\small{$\pm$0.5} & 88.2\small{$\pm$0.3} & 88.3\small{$\pm$0.3} & 88.2\small{$\pm$0.3} \\
B $\rightarrow$ A & 84.8 & 85.5$^*$\small{$\pm$0.5} & \textbf{86.0}$^*$\small{$\pm$0.3} & 85.7$^*$\small{$\pm$0.2} & 85.9$^*$\small{$\pm$0.4} & \textbf{86.0}$^*$\small{$\pm$0.2} & 85.9$^*$\small{$\pm$0.3} & 85.9$^*$\small{$\pm$0.2} \\
B $\rightarrow$ I & \textbf{83.4} & 80.1\small{$\pm$3.2} & 82.2\small{$\pm$0.7} & 82.1\small{$\pm$1.0} & 81.6\small{$\pm$1.0} & 82.2\small{$\pm$0.8} & 82.4\small{$\pm$0.5} & 82.4\small{$\pm$0.6} \\
D $\rightarrow$ B & 85.0 & 86.0\small{$\pm$1.4} & 86.8$^*$\small{$\pm$0.4} & \textbf{87.0}$^*$\small{$\pm$0.2} & \textbf{87.0}$^*$\small{$\pm$0.3} & \textbf{87.0}$^*$\small{$\pm$0.1} & \textbf{87.0}$^*$\small{$\pm$0.2} & \textbf{87.0}$^*$\small{$\pm$0.2} \\
D $\rightarrow$ E & 83.8 & 84.2\small{$\pm$1.1} & 84.9$^*$\small{$\pm$0.6} & 85.4$^*$\small{$\pm$0.3} & 85.4$^*$\small{$\pm$0.2} & 85.4$^*$\small{$\pm$0.3} & \textbf{85.6}$^*$\small{$\pm$0.3} & 85.5$^*$\small{$\pm$0.3} \\
D $\rightarrow$ K & 85.3 & \textbf{87.0}$^*$\small{$\pm$0.2} & 86.5$^*$\small{$\pm$0.6} & 86.7$^*$\small{$\pm$0.2} & 86.7$^*$\small{$\pm$0.5} & 86.8$^*$\small{$\pm$0.6} & 86.7$^*$\small{$\pm$0.5} & 86.7$^*$\small{$\pm$0.7} \\
D $\rightarrow$ A & 81.0 & 84.6$^*$\small{$\pm$0.6} & \textbf{84.8}$^*$\small{$\pm$0.6} & 84.6$^*$\small{$\pm$0.8} & 84.1$^*$\small{$\pm$0.6} & 84.3$^*$\small{$\pm$0.6} & 84.2$^*$\small{$\pm$0.2} & 84.2$^*$\small{$\pm$0.7} \\
D $\rightarrow$ I & 82.3 & \textbf{83.0}\small{$\pm$0.6} & 82.9$^*$\small{$\pm$0.5} & 82.9$^*$\small{$\pm$0.4} & 82.8\small{$\pm$0.5} & 82.8\small{$\pm$0.5} & 82.9$^*$\small{$\pm$0.1} & 82.8\small{$\pm$0.5} \\
E $\rightarrow$ B & 85.0 & 84.3\small{$\pm$0.9} & 84.6\small{$\pm$0.6} & 84.7\small{$\pm$0.3} & 84.9\small{$\pm$0.4} & \textbf{85.1}\small{$\pm$0.5} & \textbf{85.1}\small{$\pm$0.3} & 85.0\small{$\pm$0.5} \\
E $\rightarrow$ D & 84.4 & 84.3\small{$\pm$0.6} & 83.7\small{$\pm$1.6} & 84.7\small{$\pm$0.5} & \textbf{84.8}\small{$\pm$0.4} & 84.7\small{$\pm$0.2} & 84.6\small{$\pm$0.3} & \textbf{84.8}\small{$\pm$0.3} \\
E $\rightarrow$ K & 90.6 & 90.3\small{$\pm$0.3} & 90.6\small{$\pm$0.4} & 90.6\small{$\pm$0.1} & 90.7\small{$\pm$0.3} & 90.6\small{$\pm$0.4} & \textbf{90.9}$^*$\small{$\pm$0.2} & 90.7\small{$\pm$0.4} \\
E $\rightarrow$ A & 84.3 & 85.8$^*$\small{$\pm$0.9} & \textbf{86.6}$^*$\small{$\pm$0.4} & 86.3$^*$\small{$\pm$0.2} & 86.0$^*$\small{$\pm$0.7} & 86.1$^*$\small{$\pm$0.6} & 86.4$^*$\small{$\pm$0.3} & 86.1$^*$\small{$\pm$0.6} \\
E $\rightarrow$ I & 79.1 & 79.8\small{$\pm$2.1} & 80.5\small{$\pm$1.2} & 80.7$^*$\small{$\pm$0.3} & 80.9$^*$\small{$\pm$0.5} & 80.8$^*$\small{$\pm$0.3} & \textbf{81.2}$^*$\small{$\pm$0.5} & 80.9$^*$\small{$\pm$0.4} \\
K $\rightarrow$ B & 84.9 & 82.4\small{$\pm$3.2} & 82.2\small{$\pm$4.6} & 84.4\small{$\pm$1.5} & 85.0\small{$\pm$0.2} & \textbf{85.1}\small{$\pm$0.2} & 84.4\small{$\pm$1.8} & \textbf{85.1}\small{$\pm$0.3} \\
K $\rightarrow$ D & 83.1 & 83.1\small{$\pm$0.7} & 81.9\small{$\pm$2.4} & 82.5\small{$\pm$1.1} & 83.0\small{$\pm$1.2} & 82.9\small{$\pm$2.1} & \textbf{83.7}$^*$\small{$\pm$0.3} & 82.6\small{$\pm$2.2} \\
K $\rightarrow$ E & 88.1 & 88.3\small{$\pm$0.4} & 88.0\small{$\pm$0.5} & 88.0\small{$\pm$0.4} & 88.1\small{$\pm$0.4} & 88.0\small{$\pm$0.4} & 88.1\small{$\pm$0.7} & \textbf{88.4}\small{$\pm$0.3} \\
K $\rightarrow$ A & 80.4 & 85.7$^*$\small{$\pm$1.0} & 85.7$^*$\small{$\pm$0.7} & 85.8$^*$\small{$\pm$0.2} & 84.5$^*$\small{$\pm$1.7} & 85.5$^*$\small{$\pm$0.8} & \textbf{85.9}$^*$\small{$\pm$0.5} & 85.3$^*$\small{$\pm$0.6} \\
K $\rightarrow$ I & 80.2 & 77.8\small{$\pm$3.1} & 78.0\small{$\pm$4.8} & 79.4\small{$\pm$1.6} & 79.3\small{$\pm$1.5} & 80.3\small{$\pm$1.1} & \textbf{80.6}\small{$\pm$1.0} & 80.1\small{$\pm$1.8} \\
A $\rightarrow$ B & 77.2 & 78.3\small{$\pm$1.7} & 79.8\small{$\pm$2.2} & 79.8\small{$\pm$1.9} & 80.9$^*$\small{$\pm$0.6} & 80.8$^*$\small{$\pm$0.7} & \textbf{80.9}$^*$\small{$\pm$0.7} & 80.8$^*$\small{$\pm$0.6} \\
A $\rightarrow$ D & 77.7 & \textbf{79.8}\small{$\pm$2.2} & 77.5\small{$\pm$2.3} & 79.3$^*$\small{$\pm$1.3} & 79.0$^*$\small{$\pm$0.9} & 78.8$^*$\small{$\pm$0.7} & 78.9\small{$\pm$1.1} & 78.7\small{$\pm$0.7} \\
A $\rightarrow$ E & 84.3 & 84.2\small{$\pm$2.9} & 84.5\small{$\pm$1.3} & 85.1$^*$\small{$\pm$0.5} & 85.0$^*$\small{$\pm$0.4} & 84.9\small{$\pm$0.5} & \textbf{85.5}$^*$\small{$\pm$0.4} & 85.0\small{$\pm$0.5} \\
A $\rightarrow$ K & 85.0 & 87.1$^*$\small{$\pm$0.5} & 86.1$^*$\small{$\pm$0.5} & 87.0$^*$\small{$\pm$0.4} & 87.0$^*$\small{$\pm$0.3} & 87.0$^*$\small{$\pm$0.2} & \textbf{87.5}$^*$\small{$\pm$0.4} & 87.0$^*$\small{$\pm$0.3} \\
A $\rightarrow$ I & 71.2 & \textbf{77.0}$^*$\small{$\pm$0.5} & 69.8\small{$\pm$10.8} & 65.7\small{$\pm$12.8} & 70.2\small{$\pm$10.7} & 68.5\small{$\pm$9.7} & 75.3$^*$\small{$\pm$1.6} & 68.6\small{$\pm$9.6} \\
I $\rightarrow$ B & 84.5 & 86.4$^*$\small{$\pm$0.9} & \textbf{87.0}$^*$\small{$\pm$0.4} & 86.7$^*$\small{$\pm$0.2} & 86.7$^*$\small{$\pm$0.1} & 86.6$^*$\small{$\pm$0.3} & 86.6$^*$\small{$\pm$0.2} & 86.7$^*$\small{$\pm$0.3} \\
I $\rightarrow$ D & 84.8 & 85.5$^*$\small{$\pm$0.4} & \textbf{86.2}$^*$\small{$\pm$0.1} & \textbf{86.2}$^*$\small{$\pm$0.3} & 85.9$^*$\small{$\pm$0.3} & 85.9$^*$\small{$\pm$0.3} & 85.9$^*$\small{$\pm$0.2} & 85.9$^*$\small{$\pm$0.4} \\
I $\rightarrow$ E & 82.0 & \textbf{86.5}$^*$\small{$\pm$0.2} & 86.2$^*$\small{$\pm$0.2} & 86.1$^*$\small{$\pm$0.4} & 86.1$^*$\small{$\pm$0.3} & 86.1$^*$\small{$\pm$0.3} & 86.3$^*$\small{$\pm$0.3} & 86.1$^*$\small{$\pm$0.2} \\
I $\rightarrow$ K & 85.2 & 87.3$^*$\small{$\pm$0.4} & 86.7$^*$\small{$\pm$0.6} & 87.0$^*$\small{$\pm$0.2} & 87.0$^*$\small{$\pm$0.3} & 87.0$^*$\small{$\pm$0.3} & \textbf{87.4}$^*$\small{$\pm$0.1} & 87.0$^*$\small{$\pm$0.3} \\
I $\rightarrow$ A & 82.0 & \textbf{84.9}$^*$\small{$\pm$0.6} & \textbf{84.9}$^*$\small{$\pm$0.1} & 83.9$^*$\small{$\pm$0.6} & 83.8$^*$\small{$\pm$0.7} & 83.7$^*$\small{$\pm$0.6} & \textbf{84.9}$^*$\small{$\pm$0.4} & 83.9$^*$\small{$\pm$0.7} \\
\hline
Average & 83.3 & 84.2~(12$^\dagger$) & 84.1~(14$^\dagger$) & 84.3~(17$^\dagger$) & 84.4~(18$^\dagger$) & 84.4~(16$^\dagger$) & \textbf{84.9}~$(21^\dagger)$ & 84.5~$(15^\dagger)$ \\
\specialrule{.1em}{.05em}{.05em}
\end{tabular}
}
\caption{Experimental results with varying temperature values are shown. The asterisk, $^*$, denotes a value greater than the baseline with a significance level of 0.05, and $^\dagger$ represents the number of values  significantly greater than the baseline.}
\label{tab:varying temperatures}
\end{table*}

\section{Experiments}
\subsection{Experimental protocol}
In our experiments, we evaluate our approach for the task of cross-domain sentiment classification. We compare algorithms on the Airline review dataset (A)~\citep{Quang2015Airline}, IMDB dataset (I)~\citep{Mass2011IMDB}, and Amazon reviews datasets~\citep{Blitzer2007B4} which contain four domains: books (B), dvds (D), electronics (E) and Kitchen appliances (K). In total, we perform 30 domain adaptation tasks. For each domain, we sample 2,000 labeled reviews, consisting of 1,000 positive and 1,000 negative reviews. Among all the source and target examples, we use 1,600 labeled source examples and 1,600 unlabeled target examples for training, and the remaining 400 labeled source examples are used for development. Then, we finally utilize all the labeled target examples for evaluation. 

\subsection{Baselines}
Because of the powerfulness of a pre-training language model, fine-tuned BERT on a source dataset without any domain adaptation technique overwhelms the performance of existing unsupervised domain adaptation algorithms.
Accordingly, we decide to use the fine-tuned BERT, which is trained on source data only, as a basic baseline, denoting it by $\text{BERT}_\text{BASE}$. Since baseline performance is obtained by the baseline model without any domain adaptation, the primary interest is to observe whether or not the proposed model exceeds the baseline model, and comparisons with other domain adaptation methods follow.  Moreover, to see the baseline model effect, we also evaluate the model performance with the baseline models by $\text{DistilBERT}$~\citep{sanh2019distilbert} and $\text{RoBERT}$~\citep{Liu2019Roberta} models respectively. In addition, we also consider DDC~\citep{Tzeng2014DDC}, DANN~\citep{Ganin2016DANN}, and deepCORAL~\citep{Sun2016DeepCORAL} methods applied to BERT because these are designed for deep neural network architectures. For the DDC method, we use a Gaussian kernel for the MMD loss because it shows better performance than a linear kernel.

\subsection{Experimental results}
To evaluate our proposed method, we experiment with the pre-trained $\text{BERT}_{\text{BASE}}$ uncased model implemented by HuggingFace\footnote[2]{\url{https://github.com/huggingface/transformers}}~\citep{wolf2019huggingfaces}. Hyper-parameters and experimental details are described in Appendix~\ref{appendix: Experimental Details} and experimental results with $\text{DistilBERT}_\text{BASE}$  and $\text{RoBERTa}_{\text{BASE}}$ are supplemented in Appendix~\ref{appendix:Further Experiments}.

We perform five replications using different random seeds for each cross-domain pair and provide the average value and the standard deviation. We provide the results of one-sample Wilcoxon signed rank test for each algorithm as described in following tables, with baseline value as the population mean of the null hypothesis. We also perform one-sample t test, and the results are similar to those of Wilcoxon test. From Table~\ref{tab:BERT_results}, we notice that our proposed method, \emph{Adversarial Adaptation with Distillation} (AAD), significantly outperforms the baseline on 21 among 30 dataset pairs and acheives as much as 1.6\% performance gain on average over baseline. While DDC and DANN methods show worse performance than the baseline on average, the deep CORAL method shows better performance but still worse as much as 1.1\% than AAD. We notice that, though their performance improves over baseline, the existing algorithms yield lower accuracy than AAD in most cases. In addition, since our algorithm has relatively small standard deviation values, we assert that it shows more stable performance improvements.  On the contrary, other algorithms perform worse in many cases compared to that of baseline.
This is due to the ADDA framework, on which our algorithm is based, which separates fine-tuning and domain adaptation procedures. Since the BERT model has a large number of parameters, the model performance is sensitive to training-related parameters and the training scheme.

\subsection{Effect of the temperature value}
In our algorithm, knowledge distillation (KD) is a major component, and the temperature $t$ in KD determines how much the model softens the distribution. To show the effect of the temperature value, we conduct the same experiments over $t$ from \{1, 2, 5, 10, 20, 50\}. We also compare them with the supervised learning approach which can be optimized by the following objective:

\begin{align}
    \min_{E_t}~\mathcal{L}_{T}(\mathbf{X}_S, \mathbf{X}_T, \mathbf{y}_S)=\mathcal{L}_{gen}(\mathbf{X}_T) + \mathbb{E}_{(\boldsymbol{x_s}, y_s) \sim (\mathbb{X}_S, \mathbb{Y}_S)}-\sum_{k=1}^K \mathds{1}_{[k=y_s]} \log{C(E_t(\boldsymbol{x_s}))}.
    \label{eq7}
\end{align}

The results obtained by varying the value of temperature for KD and the supervised learning approach are summarized in Table~\ref{tab:varying temperatures}. Except $t=1$, the knowledge distillation method consistently outperforms the supervised learning approach both in terms of average accuracy and the number of values significantly greater than the baseline. Furthermore, the results show that as the temperature value increases up to $20$, the proposed algorithm not only has better performance but also smaller standard deviation values by and large. When $t=50$, the performance decreases because it severely dilutes correct label information. These results indicate that the knowledge distillation method with the proper temperature value $t$ 
enables the BERT model to maintain class information as well as the flexibility for adversarial adaptation as explained in section~\ref{section3.2:step2}.

\section{Conclusions and Future Work}
We presented a new method for BERT unsupervised domain adaptation which combines the ADDA framework and knowledge distillation. The direct application of ADDA to BERT resulted in catastrophic forgetting in terms of the original task. To overcome this problem, we adopted knowledge distillation and demonstrated that it can successfully integrate adversarial domain adaptation; thus the proposed method reliably improves performance on cross-domain sentiment classification tasks. We also analyzed how the algorithm works as the temperature value for knowledge distillation changes. Through the analysis, we found that a proper selection of temperature value is important since too large temperature values severely dilute correct label information, leading to performance degradation. In future, we would like to verify that our algorithm can be applied for domain adaptation in other natural language processing tasks. Our implementation is based on Pytorch~\citep{paszke2017automatic} and publicly available\footnote[3]{\url{https://github.com/bzantium/bert-AAD}\\}.

\newpage
\bibliography{main}
\bibliographystyle{main}

\appendix
\section{{Hyperparameters and Training Details}}
\label{appendix: Experimental Details}
Since we use BERT as the main component in all experiments, we follow a similar training scheme for fine-tuned BERT suggested by the original BERT paper~\citep{Devlin2018BERT}. For all the experiments with our proposed method, source BERT and classifier is trained for 3 epochs with an Adam optimizer~\citep{kingma2014adam} with a batch size of 64 and learning rate of 5e-5, $\beta_1$=0.9 and $\beta_2$=0.999 in step 1. In step 2, while target BERT and the discriminator is trained for 3 epochs with a batch size of 64 and learning rate of 1e-5, we also apply gradient clipping on the target encoder with gradient norm of 1.0 and on the discriminator with a clip value of 0.01 to make adversarial training more stable. For DDC, DANN, and deep CORAL methods, the models are trained with the same settings as step 1 for our method. In addition, we use a maximum sequence length of 128 for all tested methods.

For discriminator architecture, we use the following topology: linear(768, 3072) $\rightarrow$ leakyReLU(0.01) $\rightarrow$ linear(3072, 3072) $\rightarrow$ leakyReLU(0.01) $\rightarrow$ linear(3072, 1) $\rightarrow$ sigmoid().\footnote[4]{linear($i$, $o$) stands for a fully connected layer with a matrix of $i \times o$, and leakyReLU($\alpha$) represents the leakyReLU activation layer with negative slope $\alpha$. sigmoid() represents a sigmoid activation layer.}

\section{Further Experiments}
\label{appendix:Further Experiments}

In this section, we provide additional experimental results with baseline by $\text{DistilBERT}$ and $\text{RoBERTa}$ models summarized in Table~\ref{tab:DistilBERT_results}-\ref{tab:RoBERTa_results}. We again use HuggingFace's implementation for pre-trained models.

Similar to the results by the baseline by the BERT uncased model, our proposed method with baseline by either DistilBERT or RoBERTa shows consistently better performance than the baseline and other algorithms.

\begin{table*}[htbp]
\centering
\resizebox{10cm}{!}{
\begin{tabular}{c || lllll}
\specialrule{.1em}{.05em}{.05em}
Source $\rightarrow$ Target & Baseline & DDC & DANN & deep CORAL & AAD~(Ours) \\
\hline
B $\rightarrow$ D & 84.3 & \textbf{84.9}\small{$\pm$0.7} & 83.2\small{$\pm$1.3} & \textbf{84.9}\small{$\pm$0.7} & 84.6\small{$\pm$0.3} \\
B $\rightarrow$ E & 83.6 & 83.3\small{$\pm$1.5} & 81.5\small{$\pm$1.9} & 83.9\small{$\pm$0.9} & \textbf{84.2}$^*$\small{$\pm$0.2} \\
B $\rightarrow$ K & 84.5 & 86.4$^*$\small{$\pm$0.6} & 84.8\small{$\pm$0.9} & \textbf{86.7}$^*$\small{$\pm$0.4} & 86.1$^*$\small{$\pm$0.3} \\
B $\rightarrow$ A & 82.3 & 82.3\small{$\pm$1.5} & 80.9\small{$\pm$1.7} & 83.2\small{$\pm$1.2} & \textbf{84.8}$^*$\small{$\pm$0.3} \\
B $\rightarrow$ I & \textbf{80.9} & \textbf{80.9}\small{$\pm$1.4} & 80.2\small{$\pm$1.1} & 81.4\small{$\pm$0.9} & 80.4\small{$\pm$0.3} \\
D $\rightarrow$ B & 82.5 & \textbf{83.9}$^*$\small{$\pm$0.7} & 83.0\small{$\pm$1.9} & 84.5$^*$\small{$\pm$0.8} & 84.1$^*$\small{$\pm$0.2} \\
D $\rightarrow$ E & 80.2 & 80.1\small{$\pm$1.8} & 79.9\small{$\pm$1.8} & 81.9$^*$\small{$\pm$1.1} & \textbf{83.0}$^*$\small{$\pm$0.4} \\
D $\rightarrow$ K & 83.9 & 84.3\small{$\pm$0.8} & 83.4\small{$\pm$1.0} & 84.6\small{$\pm$0.8} & \textbf{85.2}$^*$\small{$\pm$0.2} \\
D $\rightarrow$ A & 80.5 & 79.2\small{$\pm$1.8} & 79.5\small{$\pm$2.2} & 80.6\small{$\pm$1.3} & \textbf{83.4}$^*$\small{$\pm$0.2} \\
D $\rightarrow$ I & 81.8 & 80.9\small{$\pm$0.7} & 81.0\small{$\pm$1.2} & 81.1\small{$\pm$0.7} & \textbf{82.2}$^*$\small{$\pm$0.2} \\
E $\rightarrow$ B & 83.3 & 81.8\small{$\pm$2.3} & 82.9\small{$\pm$0.8} & 82.5\small{$\pm$1.8} & \textbf{84.5}$^*$\small{$\pm$0.1} \\
E $\rightarrow$ D & 82.7 & 81.8\small{$\pm$0.9} & 82.5\small{$\pm$0.6} & 82.4\small{$\pm$0.9} & \textbf{83.0}$^*$\small{$\pm$0.2} \\
E $\rightarrow$ K & 88.2 & 87.3\small{$\pm$1.6} & 88.2\small{$\pm$0.6} & 87.6\small{$\pm$1.4} & \textbf{88.8}$^*$\small{$\pm$0.2} \\
E $\rightarrow$ A & 83.9 & \textbf{84.7}$^*$\small{$\pm$0.4} & 84.6\small{$\pm$0.9} & 85.0$^*$\small{$\pm$0.2} & 84.6$^*$\small{$\pm$0.2} \\
E $\rightarrow$ I & 78.3 & 78.4\small{$\pm$0.4} & 78.7\small{$\pm$1.1} & 79.1$^*$\small{$\pm$0.4} & \textbf{79.4}$^*$\small{$\pm$0.2} \\
K $\rightarrow$ B & 82.2 & 79.3\small{$\pm$2.8} & 81.4\small{$\pm$1.6} & 81.7\small{$\pm$2.4} & \textbf{84.3}$^*$\small{$\pm$0.3} \\
K $\rightarrow$ D & 82.6 & 81.5\small{$\pm$1.1} & 81.4\small{$\pm$0.8} & 81.8\small{$\pm$1.2} & \textbf{83.0}$^*$\small{$\pm$0.3} \\
K $\rightarrow$ E & \textbf{87.4} & 86.3\small{$\pm$0.9} & 86.1\small{$\pm$1.3} & 86.3\small{$\pm$0.8} & \textbf{87.4}\small{$\pm$0.1} \\
K $\rightarrow$ A & 83.4 & 81.4\small{$\pm$2.3} & 81.3\small{$\pm$1.8} & 83.3\small{$\pm$1.5} & \textbf{85.3}$^*$\small{$\pm$0.1} \\
K $\rightarrow$ I & \textbf{79.8} & 78.3\small{$\pm$0.5} & 78.2\small{$\pm$0.9} & 78.7\small{$\pm$0.8} & 79.2\small{$\pm$0.2} \\
A $\rightarrow$ B & 76.9 & 77.9\small{$\pm$1.4} & 74.8\small{$\pm$5.4} & 79.4$^*$\small{$\pm$1.0} & \textbf{80.1}$^*$\small{$\pm$0.2} \\
A $\rightarrow$ D & 76.6 & 77.8\small{$\pm$1.5} & 75.6\small{$\pm$4.9} & 78.9$^*$\small{$\pm$1.1} & \textbf{79.0}$^*$\small{$\pm$0.3} \\
A $\rightarrow$ E & 82.8 & 82.8\small{$\pm$0.7} & 81.2\small{$\pm$2.1} & \textbf{83.4}$^*$\small{$\pm$0.7} & 82.8\small{$\pm$0.1} \\
A $\rightarrow$ K & 83.6 & 83.4\small{$\pm$0.2} & 81.3\small{$\pm$4.6} & \textbf{84.4}$^*$\small{$\pm$0.5} & 83.3\small{$\pm$0.3} \\
A $\rightarrow$ I & 71.2 & 73.6\small{$\pm$2.1} & 71.8\small{$\pm$5.3} & \textbf{75.4}$^*$\small{$\pm$1.0} & 71.0\small{$\pm$2.5} \\
I $\rightarrow$ B & 83.3 & 83.1\small{$\pm$1.1} & 81.1\small{$\pm$1.9} & 83.8\small{$\pm$0.7} & \textbf{84.8}$^*$\small{$\pm$0.3} \\
I $\rightarrow$ D & 84.9 & 84.5\small{$\pm$0.1} & 82.9\small{$\pm$1.0} & 84.6\small{$\pm$0.2} & \textbf{86.1}$^*$\small{$\pm$0.1} \\
I $\rightarrow$ E & 80.5 & 81.5$^*$\small{$\pm$1.1} & 81.8\small{$\pm$1.3} & 82.8$^*$\small{$\pm$1.1} & \textbf{84.5}$^*$\small{$\pm$0.4} \\
I $\rightarrow$ K & 83.9 & 84.5\small{$\pm$0.9} & 83.8\small{$\pm$1.5} & 84.8$^*$\small{$\pm$0.8} & \textbf{85.0}$^*$\small{$\pm$0.3} \\
I $\rightarrow$ A & 77.7 & 81.0$^*$\small{$\pm$1.2} & 82.1$^*$\small{$\pm$1.3} & 82.3$^*$\small{$\pm$0.9} & \textbf{83.1}$^*$\small{$\pm$2.5} \\
\hline
Average & 81.9 & 81.9~(5$^\dagger$) & 81.3~(1$^\dagger$) & 82.7~(13$^\dagger$) & \textbf{83.2}~(23$^\dagger$) \\
\specialrule{.1em}{.05em}{.05em}
\end{tabular}
}
\caption{Sentiment classification accuracy with $\text{DistilBERT}_\text{BASE}$ and other tested models including the proposed model ADD on 30 cross-domain sentiment classification tasks is shown. The asterisk, $^*$, denotes that a value greater than the baseline with a significance level of 0.05, and $^\dagger$ represents the number of values significantly greater than the baseline.}
\label{tab:DistilBERT_results}
\end{table*}

\begin{table*}[!htbp]
\centering
\resizebox{10cm}{!}{
\begin{tabular}{c || lllll}
\specialrule{.1em}{.05em}{.05em}
Source $\rightarrow$ Target & Baseline & DDC & DANN & deep CORAL & AAD~(Ours) \\
\hline
B $\rightarrow$ D & 88.4 & 88.2\small{$\pm$0.5} & 87.4\small{$\pm$1.0} & 87.1\small{$\pm$0.9} & \textbf{88.5}\small{$\pm$0.2} \\
B $\rightarrow$ E & 90.0 & 89.5\small{$\pm$0.8} & 88.1\small{$\pm$1.3} & 88.9\small{$\pm$1.8} & \textbf{90.1}\small{$\pm$0.3} \\
B $\rightarrow$ K & 90.9 & 90.4\small{$\pm$1.6} & 90.6\small{$\pm$1.6} & 90.3\small{$\pm$2.0} & \textbf{92.2}$^*$\small{$\pm$0.1} \\
B $\rightarrow$ A & 85.4 & 85.0\small{$\pm$1.3} & 83.3\small{$\pm$2.1} & 85.1\small{$\pm$2.3} & \textbf{87.0}$^*$\small{$\pm$0.4} \\
B $\rightarrow$ I & \textbf{86.3} & 86.1\small{$\pm$0.6} & \textbf{86.3}\small{$\pm$0.8} & 85.4\small{$\pm$0.7} & 86.2\small{$\pm$0.2} \\
D $\rightarrow$ B & 87.4 & 89.6$^*$\small{$\pm$0.9} & 88.1\small{$\pm$1.6} & 88.9\small{$\pm$1.5} & \textbf{89.6}$^*$\small{$\pm$0.1} \\
D $\rightarrow$ E & 89.9 & 86.4\small{$\pm$3.5} & 85.4\small{$\pm$4.2} & 87.9\small{$\pm$3.8} & \textbf{90.2}$^*$\small{$\pm$0.2} \\
D $\rightarrow$ K & 88.9 & 89.6\small{$\pm$1.8} & 88.0\small{$\pm$1.5} & 90.0\small{$\pm$0.9} & \textbf{91.5}$^*$\small{$\pm$0.3} \\
D $\rightarrow$ A & 85.3 & 82.8\small{$\pm$4.7} & 84.0\small{$\pm$1.5} & 84.5\small{$\pm$1.3} & \textbf{86.7}$^*$\small{$\pm$0.2} \\
D $\rightarrow$ I & 87.2 & 85.8\small{$\pm$0.8} & 86.1\small{$\pm$0.7} & 85.9\small{$\pm$2.1} & \textbf{87.4}\small{$\pm$0.3} \\
E $\rightarrow$ B & 85.4 & 86.7\small{$\pm$1.4} & 86.5\small{$\pm$0.9} & 88.2\small{$\pm$1.6} & \textbf{88.6}$^*$\small{$\pm$0.3} \\
E $\rightarrow$ D & 83.8 & 85.3\small{$\pm$1.8} & 84.1\small{$\pm$1.6} & 86.0$^*$\small{$\pm$0.6} & \textbf{87.4}$^*$\small{$\pm$0.6} \\
E $\rightarrow$ K & 92.5 & 92.0\small{$\pm$1.0} & 91.2\small{$\pm$1.8} & 91.9\small{$\pm$0.7} & \textbf{93.1}$^*$\small{$\pm$0.1} \\
E $\rightarrow$ A & 83.9 & 86.7$^*$\small{$\pm$0.7} & 86.5$^*$\small{$\pm$0.9} & 86.8$^*$\small{$\pm$1.5} & \textbf{87.2}$^*$\small{$\pm$0.6} \\
E $\rightarrow$ I & 78.9 & 82.2\small{$\pm$3.2} & 82.2$^*$\small{$\pm$1.4} & 83.1$^*$\small{$\pm$1.3} & \textbf{85.2}$^*$\small{$\pm$0.1} \\
K $\rightarrow$ B & 87.7 & 88.7$^*$\small{$\pm$0.8} & 86.3\small{$\pm$1.3} & 88.0\small{$\pm$0.9} & \textbf{88.9}$^*$\small{$\pm$0.3} \\
K $\rightarrow$ D & \textbf{86.7} & 85.8\small{$\pm$1.8} & 84.7\small{$\pm$1.9} & 86.3\small{$\pm$0.7} & 86.1\small{$\pm$0.3} \\
K $\rightarrow$ E & 90.2 & 89.0\small{$\pm$1.8} & 90.0\small{$\pm$1.4} & 90.3\small{$\pm$1.2} & \textbf{90.9}$^*$\small{$\pm$0.2} \\
K $\rightarrow$ A & 86.4 & 82.9\small{$\pm$2.5} & 84.6\small{$\pm$3.2} & 85.4\small{$\pm$1.1} & \textbf{86.6}\small{$\pm$0.2} \\
K $\rightarrow$ I & \textbf{85.1} & 83.1\small{$\pm$2.1} & 84.0\small{$\pm$2.0} & 84.7\small{$\pm$1.0} & 85.0\small{$\pm$0.2} \\
A $\rightarrow$ B & \textbf{84.1} & 80.1\small{$\pm$4.1} & 79.9\small{$\pm$3.1} & 84.5\small{$\pm$3.6} & 82.8\small{$\pm$1.6} \\
A $\rightarrow$ D & 83.2 & 79.9\small{$\pm$3.0} & 81.4\small{$\pm$2.0} & 83.2\small{$\pm$0.6} & \textbf{84.2}$^*$\small{$\pm$0.6} \\
A $\rightarrow$ E & 86.9 & 86.1\small{$\pm$2.7} & 86.6\small{$\pm$3.0} & \textbf{88.0}\small{$\pm$2.8} & 87.8\small{$\pm$0.8} \\
A $\rightarrow$ K & 87.2 & 85.4\small{$\pm$3.6} & 88.4$^*$\small{$\pm$1.2} & 86.3\small{$\pm$4.5} & \textbf{89.7}$^*$\small{$\pm$0.5} \\
A $\rightarrow$ I & 81.9 & 77.7\small{$\pm$5.3} & 77.4\small{$\pm$3.0} & 82.0\small{$\pm$1.9} & \textbf{82.3}\small{$\pm$0.6} \\
I $\rightarrow$ B & \textbf{90.0} & 86.6\small{$\pm$2.1} & 86.2\small{$\pm$4.0} & 89.6\small{$\pm$0.9} & 89.9\small{$\pm$0.3} \\
I $\rightarrow$ D & 87.4 & 86.8\small{$\pm$1.1} & 87.4\small{$\pm$0.7} & 86.1\small{$\pm$2.8} & \textbf{88.7}$^*$\small{$\pm$0.1} \\
I $\rightarrow$ E & 88.6 & 85.0\small{$\pm$4.0} & 85.9\small{$\pm$2.2} & 88.2\small{$\pm$1.8} & \textbf{89.5}$^*$\small{$\pm$0.3} \\
I $\rightarrow$ K & \textbf{90.8} & 85.4\small{$\pm$5.1} & 86.6\small{$\pm$2.7} & 90.6\small{$\pm$1.2} & 90.6\small{$\pm$0.3} \\
I $\rightarrow$ A & 85.7 & 84.6\small{$\pm$2.2} & 84.5\small{$\pm$1.9} & 86.1\small{$\pm$0.8} & \textbf{86.1}\small{$\pm$0.8} \\
\hline
Average & 86.8 & 85.8~(3$^\dagger$) & 85.7~(3$^\dagger$) & 87.0~(3$^\dagger$) & \textbf{88.0}~(17$^\dagger$) \\
\specialrule{.1em}{.05em}{.05em}
\end{tabular}
}
\caption{Sentiment classification accuracy with $\text{RoBERTa}_\text{BASE}$ and other tested models including the proposed model ADD on 30 cross-domain sentiment classification tasks is shown. The asterisk, $^*$, denotes a value greater than the baseline with a significance level of 0.05, and $^\dagger$ represents the number of values significantly greater than the baseline.}
\label{tab:RoBERTa_results}
\end{table*}
\end{document}